# Robust Tracking Control for Constrained Robots

Haifa Mehdi, Olfa Boubaker*

*National Institute of Applied Sciences and Technology, Centre Urbain Nord, BP 676 - 1080 Tunis, Tunisia*

**Abstract**

In this paper, a novel robust tracking control law is proposed for constrained robots under unknown stiffness environment. The stability and the robustness of the controller are proved using a Lyapunov-based approach where the relationship between the error dynamics of the robotic system and its energy is investigated. Finally, a 3DOF constrained robotic arm is used to prove the stability, the robustness and the safety of the proposed approach.

*Keywords*: Position/force control; Robust control; Unknown stiffness environment; Constrained robot.

## 1. Introduction

Robust tracking control under unknown constrained environment is one of the main important issues in robotic field. A number of significant papers are proposed in this area. They include fuzzy logic based controllers [1, 2], neural networks based controllers [3, 4] and sliding mode controllers [5, 6]. In the most cases, these research papers are not concerned by the compromise between robustness and safety for the tracking control problem. Only few papers are devoted to such subject [7, 8]. This paper is placed in this context. Furthermore, the controller proposed in this paper is simpler than those proposed in [7] and [8]. The stability and the robustness of the proposed approach are shown, by simulation results, on a robotic manipulator constrained to a circular trajectory.

## 2. Problem formulation

Consider a constrained robotic system with $n$ degrees of freedom described by the dynamical model [9]:

$$M(\theta)\ddot{\theta} + H(\theta,\dot{\theta}) + G(\theta) = U - J(\theta)^T F \quad (1)$$

where $\theta, \dot{\theta}, \ddot{\theta} \in R^n$ are joint position, joint velocity and joint acceleration vectors, respectively. $M(\theta) \in R^{n \times n}$ is the inertia matrix, $H(\theta,\dot{\theta}) \in R^n$ is the vector of centrifugal and Coriolis forces and $G(\theta) \in R^n$ is the vector of gravity terms. $U \in R^n$ is the generalized joint force vector, $F \in R^p$ is the vector of contact generalized forces exerted by the manipulator on the environment and $p$ is the task space dimension. Let $X, \dot{X} \in R^p$ the Cartesian position vector and the Cartesian velocity vectors of the robotic system defined, respectively, in the task space by:

---

* Corresponding author. Tel.:+0-216-1-703-929; fax: .:+0-216-1-704-329.
*E-mail address*: olfa.boubaker@insat.rnu.tn.

$$X = h(\theta) \tag{2}$$

$$\dot{X} = \partial h(\theta)/\partial \theta = J(\theta)\dot{\theta} \tag{3}$$

where $h(\theta): R^n \to R^p$ is a vector of nonlinear functions describing the forward kinematic model and $J(\theta) \in R^{p \times n}$ is the Jacobian matrix assumed to be full ranked. The temporal derivative of the Kinematic model (3) is given by:

$$\ddot{X} = \dot{J}(\theta)\dot{\theta} + J(\theta)\ddot{\theta} \tag{4}$$

where $\ddot{X} \in R^p$ is Cartesian end-effector acceleration.

Given a desired Cartesian position of the end-effector $X_d \in R^p$ the control problem aims to ensure:

$$\lim_{t \to t_f} X_d - X = 0 \tag{5}$$

**Notation:** In the following, we will adopt the following notations:

$$M(\theta) = M, \quad H(\theta, \dot{\theta}) = H, \quad G(\theta) = G, \quad J(\theta) = J$$

## 3. Robust position/force controller

**Theorem:**

The constrained robotic system described by the dynamical model:

$$M\ddot{\theta} + H + G = U - J^T F \tag{6}$$

is asymptotically stable under the uncertain force model:

$$F = (K_e + \Delta K_e)(X_d - X) \tag{7}$$

where $K_e$ and $\Delta K_e \in R^{p \times p}$ are the environment stiffness and the stiffness unknown uncertainty, respectively, and the robust control law is described by:

$$U = J^T[K_p(X_d - X) + K_v(\dot{X}_d - \dot{X})] + J^T(\dot{X}_d - \dot{X})K\sigma\, sign(\sigma) + G \tag{8}$$

where $\sigma$ is a nonlinear function defined by:

$$\sigma = C\left[(\dot{X}_d - \dot{X}) + \Lambda(X_d - X)\right] \tag{9}$$

for the constant vector $C \in R^{1 \times p}$ and the positive constant matrices $K$ and $\Lambda \in R^{p \times p}$, if there exist diagonal gain matrices $K_p, K_v \in R^{p \times p}$ satisfying the following conditions:

$$\begin{cases} \Delta K_e - K_p + K_e < 0 \\ K_v > 0 \end{cases} \tag{10}$$

**Remark:**

To satisfy safety of the robotic system and the environment, controller (8) is based on the uncertain force model (7). The nonlinear function $sign(\sigma)$ is used to ensure robustness by controlling at the same time constrained robot motion and constraint force. On the other hand, the nonlinear function is also able to limit the degradation of tracking performance occurring during saturation.

**Stability Proof:**

Consider for the constrained robot system (6) the error vector defined in the joint space by:

$$\Phi = \theta - \theta_d \tag{11}$$

and the error vector defined in the task space by:

$$Y(\Phi) = X(\theta) - X_d \tag{12}$$

Under the unknown force model (7) and the robust control law (8) we can write that:

$$M(\Phi)\ddot{\Phi} + H(\Phi,\dot{\Phi}) + J^T(\Phi)K_1 Y(\Phi) + J^T(\Phi)K_2\dot{Y}(\Phi) + J^T(\Phi)\dot{Y}(\Phi)K\sigma\, sign(\sigma) = 0 \tag{13}$$

where :

$$K_1 = K_p - K_e - \Delta K_e$$
$$K_2 = K_v$$

Following the some method presented in [10, 11] we can prove that:

$$H(\Phi,\dot{\Phi}) = \sum_{i=1}^{n}\left(\frac{d\Phi_i}{dt}\frac{\partial M(\Phi)}{\partial \Phi_i}\right)\frac{\dot{\Phi}}{2} \tag{14}$$

$$\frac{\partial P(\Phi)}{\partial \Phi} = J^T(\Phi)K_1 Y(\Phi) \tag{15}$$

$$\frac{\partial D(\Phi,\dot{\Phi})}{\partial \dot{\Phi}} = J^T(\Phi)K_2\dot{Y}(\Phi) + J^T(\Phi)\dot{Y}(\Phi)K\sigma\, sign(\sigma) \tag{16}$$

Impose, now, to the system (13) to have a Lyapunov Hamiltonian function defined by:

$$V(\Phi,\dot{\Phi}) = T(\Phi,\dot{\Phi}) + P(\Phi) - P(0) \tag{17}$$

The error system (13) is asymptotically stable if $V(\Phi,\dot{\Phi})$ satisfies the three conditions imposed by Lyapunov theorem [12]. For proving the first and the second conditions, we derive the same developments as those presented in [10]. The conditions $\Delta K_e - K_p + K_e < 0$ are then well obtained. To prove the third Lyapunov condition, the derivative of the expression (17) gives:

$$\frac{dV(\Phi)}{dt} = \frac{dT(\Phi,\dot{\Phi})}{dt} + \frac{dP(\Phi)}{dt} \tag{18}$$

From equations (15) and (16) we can write:

$$\frac{dT(\Phi,\dot{\Phi})}{dt} = \dot{\Phi}^T M(\Phi)\ddot{\Phi} + \dot{\Phi}^T \frac{dM(\Phi)}{dt}\frac{\dot{\Phi}}{2} = \dot{\Phi}^T M(\Phi)\ddot{\Phi} + \dot{\Phi}^T H(\Phi,\dot{\Phi}) \quad (19)$$

Furthermore,

$$\frac{dP(\Phi)}{dt} = \dot{\Phi}^T \frac{\partial P(\Phi)}{\partial \Phi} = \dot{\Phi}^T J^T(\Phi) K_1 Y(\Phi) \quad (20)$$

So, we can prove that:

$$\frac{dV(\Phi,\dot{\Phi})}{dt} = \dot{\Phi}^T M(\Phi)\ddot{\Phi} + \dot{\Phi}^T H(\Phi,\dot{\Phi}) + \dot{\Phi}^T J^T(\Phi) K_1 Y(\Phi) \quad (21)$$

From (13) we can write:

$$M(\Phi)\ddot{\Phi} + H(\Phi,\dot{\Phi}) + J^T(\Phi)K_1 Y(\Phi) = -J^T(\Phi)K_2 \dot{Y}(\Phi) - J^T(\Phi)\dot{Y}(\Phi)k\sigma\, sign(\sigma) \quad (22)$$

Substituting the second member of (22) in (21) gives:

$$\frac{dV(\Phi,\dot{\Phi})}{dt} = -\dot{\Phi}^T J^T(\Phi) K_2 \dot{Y}(\Phi) - \dot{\Phi}^T J^T(\Phi)\dot{Y}(\Phi)K\sigma\, sign(\sigma) \quad (23)$$

Using relations (3), (11) and (12) gives:

$$\frac{dV(\Phi,\dot{\Phi})}{dt} = -\dot{Y}^T(\Phi) K_2 \dot{Y}(\Phi) - \dot{Y}^T(\Phi)\dot{Y}(\Phi)K\sigma\, sign(\sigma) \quad (24)$$

The third Lyapunov condition is then verified if $K_v$ is positive definite.

## 4. Simulation results

Simulation results are carried out using a 3DOF robotic system using the physical parameter data given in [13, 14] for the constrained circular motion described by:

$$\begin{cases} x_d(t) = 0.76\cos(3\pi t^2 - 2\pi t^3) \\ y_d(t) = 0.76\sin(3\pi t^2 - 2\pi t^3) \end{cases}$$

for $\theta_{i0} = [0\ \ 0\ \ 0]^T$ $\theta_{id} = [\pi\ \ \pi\ \ \pi]^T$, $t_0 = 0$ and $t_f = 1s$. For sufficient conditions (10), the numerical parameters are chosen as:

$$\Lambda = diag[10\ \ 10]\ K = 30, C = [1\ \ 1]\ K_e = diag[100\ \ 100],$$
$$\Delta K_e = diag[20\ \ 50],\ K_p = diag[300\ \ 500],\ K_v = diag[200\ \ 350].$$

Fig.1 and Fig. 2 show the evolution of the robot in the Cartesian space with respect to the constrained circular trajectory and the smooth profile of the robust control laws (8), respectively.

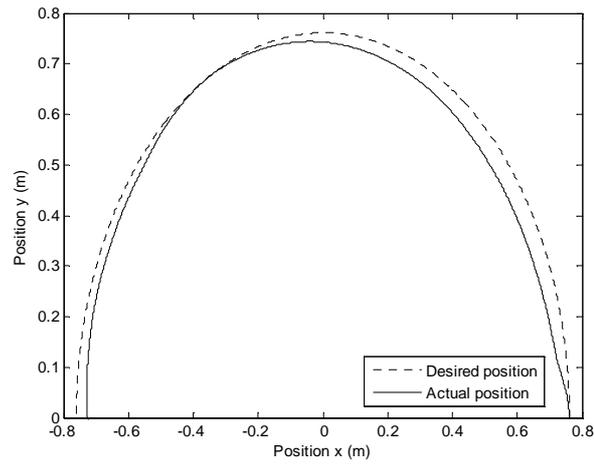

Fig. 1. End effector trajectory

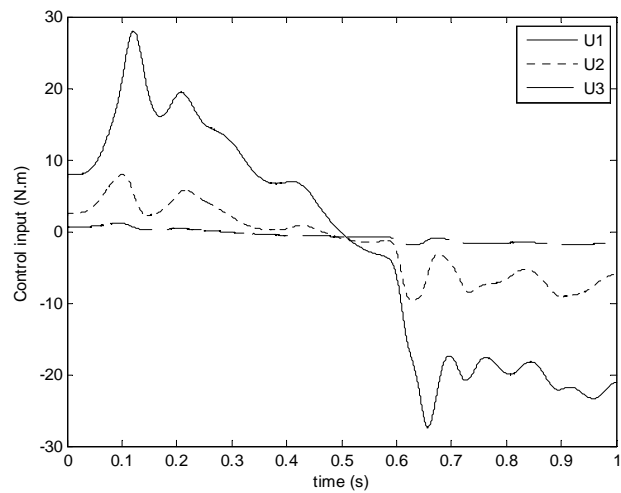

Fig. 2. Control laws

**Discussion:**

It 'is clear that the current control algorithm can be implemented in many real applications. For example, the proposed approach can be relevant for rehabilitation device applications, see for example [10-11], and for bipedal and humanoid robots during the impact and double support phases, see for example [15-16]. In future investigations, a comparative analysis with the related works [7-8] are also planned.

## 5. Conclusion

This paper proposes a simple robust controller for motion tracking of constrained robots under unknown stiffness environment. The proposed approach takes care on the compromise between robustness and safety for the tracking control problem. The stability and the robustness of the controller are proved using a Lyapunov-based approach. Future investigations will concern the application of the proposed approach for humanoid and rehabilitation robotic devices.